\documentclass{article}

\PassOptionsToPackage{numbers, compress}{natbib}
\usepackage[preprint]{neurips_2021}

\usepackage[utf8]{inputenc} 
\usepackage[T1]{fontenc}    
\usepackage{hyperref}       
\usepackage{url}            
\usepackage{booktabs}       
\usepackage{amsfonts}       
\usepackage{nicefrac}       
\usepackage{microtype}      
\usepackage{xcolor}         
\usepackage{graphicx}
\usepackage{amsmath}
\usepackage{mathrsfs}
\usepackage{multirow}
\usepackage{amsfonts}
\usepackage{algorithm}
\usepackage{algorithmic}
\usepackage{pdfpages}
\usepackage{subfigure}

\newcommand{\tabincell}[2]{\begin{tabular}{@{}#1@{}}#2\end{tabular}}  

\title{Dominant Patterns: Critical Features Hidden in Deep Neural Networks}

\author{%
  Zhixing Ye$^{1}$, ~~Shaofei Qin$^{2}$, ~~Sizhe Chen$^{1}$, ~~Xiaolin Huang$^{1*}$\\
  $^1$Department of Automation, Shanghai Jiao Tong University. \\
  $^2$Department of Electrical Engineering, Shanghai Jiao Tong University.\\
  \texttt{yzx1213@sjtu.edu.cn, ~~sufer\_qin@sjtu.edu.cn~~~} \\

}

\begin{document}

\maketitle

\begin{abstract}
In this paper, we find the existence of critical features hidden in Deep Neural Networks (DNNs), which are imperceptible but can actually dominate the output of DNNs. We call these features \textit{dominant patterns}. 
As the name suggests, for a natural image, if we add the dominant pattern of a DNN to it, the output of this DNN is determined by the dominant pattern instead of the original image, i.e., DNN's prediction is the same with the dominant pattern's. 
We design an algorithm to find such patterns by pursuing the insensitivity in the feature space. A direct application of the dominant patterns is the Universal Adversarial Perturbations (UAPs). Numerical experiments show that the found dominant patterns defeat state-of-the-art UAP methods, especially in label-free settings. In addition, dominant patterns are proved to have the potential to attack downstream tasks in which DNNs share the same backbone. We claim that DNN-specific dominant patterns reveal some essential properties of a DNN and are of great importance for its feature analysis and robustness enhancement.
\end{abstract}
\section{Introduction}\label{sec:intro}
Deep Neural Networks (DNNs) are proved as susceptible to adversarial perturbations \cite{bfgs, fgsm, pgd, deepfool, cw, onepixel}.
The basic procedure of attacking a well-trained DNN $f$ is to generate an imperceptible perturbation $\boldsymbol{\delta}$ from an image $\boldsymbol{x}$, such that the perturbation could change the prediction of $f$. Mathematically, the perturbation is dependent on $\boldsymbol{x}$ and $f$, which could be denoted as $\boldsymbol{\delta} (\boldsymbol{x},f)$, and satisfies
\begin{equation}\label{attack}
f(\boldsymbol{x}+\boldsymbol{\delta} (\boldsymbol{x},f))\neq f(\boldsymbol{x}) \mathrm{~and~} \|\boldsymbol{\delta} (\boldsymbol{x},f)\|_p \leq \xi,
\end{equation}
where the latter constraint ensures that the perturbation is imperceptible by restricting it in an $\ell_p$ norm bound $\xi$. 
By contrast, the goal of defenders (also network trainers) is to train a network insensitive to any imperceptible perturbations, that
\begin{equation}
    f(\boldsymbol{x}+\boldsymbol{\delta})=f(\boldsymbol{x}), ~~\forall \|\boldsymbol{\delta}\|_p \le \xi ,~\boldsymbol{x} \in X,
\label{equ:robustdnn}
\end{equation}
where $X$ is the dataset that $f$ works on.

Typically, $\boldsymbol{\delta}$ is regarded as a special noise, which has seldomly been treated as a meaningful signal. However, if we drop the prejudice that "the images are signal, while the perturbations are noise", we will discover that the output of the DNN is actually determined by $\boldsymbol{x}$ and $\boldsymbol{\delta}$ jointly. In other words, although the amplitude of $\boldsymbol{\delta}$ is too small to be perceived by human beings, it could still be meaningful and contain much information from the perspective of $f$, and sometimes even has a more significant effect compared to $\boldsymbol{x}$. With this idea, corresponding to equation (\ref{equ:robustdnn}), we come up with a question:
Given an $f$, is there a perturbation $\boldsymbol{\delta}$ that can provoke a high response from $f$ and make it insensitive to natural images $\boldsymbol{x}$? 
That is, for a well-trained network $f$, can we find a $\boldsymbol{\delta}(f)$, satisfying
\begin{equation}\label{eq:concept}
   f(\boldsymbol{\delta}(f)+\boldsymbol{x}) =  f(\boldsymbol{\delta}(f) ) \mathrm{~and~} \|\boldsymbol{\delta} (f)\|_p \leq \xi, ~\forall \boldsymbol{x} \in X.
\end{equation}


\begin{figure}[t]
    \centering
    \includegraphics[width=.95\textwidth]{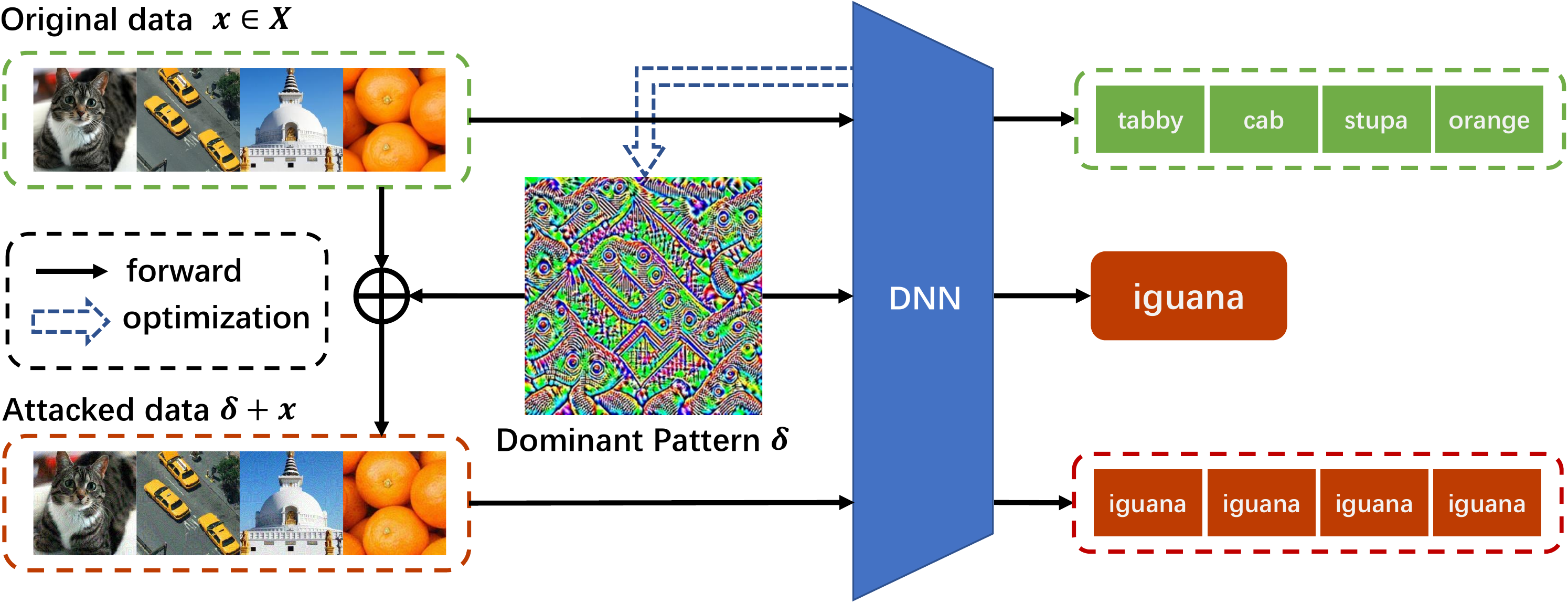}
    \caption{An example of dominant pattern. The dominant pattern is generated from a given DNN and has been mapped from [-10,10] to [0,255] for better observation. }
    \label{fig:intro}
    \vspace{-4mm}
\end{figure}
To some extent, equation (\ref{eq:concept}) can be regarded as a "dual" version of equation (\ref{equ:robustdnn}). 
For a DNN $f$, it is actually pursuing a $\boldsymbol{\delta}(f)$ that can dominate DNN's output and thus we name it as \textit{dominant pattern}. The dominant pattern is small and could be ignored in magnitude but is a critical feature hidden in DNN. Theorically, dominant patterns are data-independent and DNN-specific which could reflect some essential properties of DNN. Therefore it is a very valuable things and is helpful for network analysis, feature analysis, and perhaps robustness enhancement.



In this paper, we design an algorithm to find the dominant pattern for a given DNN. The crucial idea is to keep the output insensitive to $\boldsymbol{x} \in X$, so that make $\boldsymbol{\delta}$ more relevant to the nature of DNN . 
Here, we achieve such insensitivity by optimizing a designed Cosine-Similarity Loss (CoSLoss) to encourage the similarity between $f(\boldsymbol{\delta})$ and $f(\boldsymbol{\delta}+\boldsymbol{x})$.

As an example, we demonstrate $\boldsymbol{\delta}(f)$ and its influence on the DNN in classification task in Figure \ref{fig:intro}, where $\boldsymbol{\delta}$ is the dominant pattern found in a well-trained VGG19 \cite{vgg} model on ImageNet. The pattern is predicted as "iguana", and added with it, the original images are amazingly all mispredicted as "iguana" by VGG19.

A direct application of dominant patterns is the universal attacks. Such attacks 
generate a set of adversarial examples by crafting only one perturbation, named the Universal Adversarial Perturbation (UAP), which could be realized by a dominant pattern. Comprehensive experiments show that dominant patterns outperform state-of-the-art (SOTA) UAP methods.
Actually, a UAP pursues a perturbation to affect the output, i.e., to let the output be a function of both $\boldsymbol{x}$ and $\boldsymbol{\delta}$, while the dominant pattern pursues to determine the output, i.e., the output only depends on $\boldsymbol{\delta}$. This essential difference between them accounts for the gap in their performance.

Notably, the dominant pattern is a unique nature of DNNs instead of that for specific data. On one hand, the dominant pattern can be calculated from any unlabeled image and perform significantly better than previous label-free UAP methods. On the other hand, 
any downstream task that uses the DNN's backbone, i.e., the main part of this DNN, will also be controlled by the dominant pattern. 
The special properties of dominant patterns make them intrinsic and critical features hidden in DNNs, and provide a new viewpoint for understanding DNN's vulnerability and enhancing their robustness in the future. 

The major contributions of our work can be summarized as:
\begin{itemize}
    \item We raise an idea that a well-trained DNN can be controlled by a special perturbation named dominant pattern. We verify the existence of dominant patterns by designing an algorithm based on cosine similarity to pursue insensitivity in the feature space. 
    \item Since dominant patterns are independent of data, any unlabeled natural image set can be used in our algorithm. We substantiate that when dominant patterns are used as UAPs, they can outperform several SOTA methods, especially in label-free settings.
    \item We verify the dominance of these patterns when transferred to downstream tasks, indicating that dominant patterns are actually intrinsic features hidden in DNNs as a potential helper for model analysis and robustness enhancement.
\end{itemize}

\section{Related Work}
Recent studies have shown that DNNs are susceptible to maliciously constructed small noise called adversarial perturbations. In this section, we roughly classify them as image-specific attacks and image-agnostic ones as described below.
\subsection{Image-specific attacks}
Szegedy \textit{et al. }\cite{bfgs} first demonstrates the existence of small perturbations that can fool DNNs. 
Since then, many 
attacks have been proposed, including Fast Gradient Sign Method (FGSM) \cite{fgsm}, Projected Gradient Descent (PGD) \cite{pgd}, C\&W \cite{cw}, and DeepFool \cite{deepfool}. 
The general attack procedure is to 
optimize the input sample to adversarially maximize the loss, while restricting the magnitude of perturbations. It is also demonstrated that the adversarial examples generated by attacking a DNN may transfer to other DNNs as well \cite{dong2018boosting, xie2019improving, aoa}. 
These attacks are image-specific because they craft different perturbations for different samples. 

\subsection{Image-agnostic attacks}
Image-agnostic attacks, in contrast, change a DNN's predictions on different inputs with a single imperceptible universal adversarial perturbation (UAP). The first UAP is proposed by Moosavi-Dezfooli\textit{ et al.} \cite{uap}, 
where an iterative procedure based on Deepfool \cite{deepfool} is designed. Mopuri \textit{et al.} \cite{nag} propose a Network for Adversary Generation (NAG) to model the distribution of adversarial perturbations. Omid Poursaeed \textit{et al.} \cite{gap} present a framework named GAP which can craft universal perturbations for both classification and semantic segmentation tasks. 
Different to previous methods, GAP can achieve targeted attacks.  
The methods mentioned above require the knowledge of original training data, which may not be feasible in practice. Thus, some researchers try to train universal perturbations without using any training data \cite{fff, gduap, aaa, pdua}. In \cite{fff} and \cite{gduap}, the perturbations are generated via corrupting the extracted features at multiple layers. By contrast, \cite{aaa} trains a generative model to craft UAPs by utilizing the class impressions, which is a generic representation of the samples belonging to certain categories, while \cite{pdua}  considers the model uncertainty to craft an insensitive universal perturbation. Recently, more types of UAPs have come out to fit different demands. Philipp Benz \textit{et al.} \cite{cduap} propose a framework tailored for the class discriminative universal attack, where the perturbation fools a target network to misclassify only a chosen group of classes. Targeted UAPs can also be found by exploiting a proxy dataset instead of the original training data \cite{tuap}. 

\section{Discovering the dominant pattern}
\label{sec:method}
In this section, we formally define dominant patterns and then propose an algorithm to find them. 


As the name suggests, the dominant pattern is a small perturbation $\boldsymbol{\delta}$ that can determine the output feature $f(\boldsymbol{\delta}+\boldsymbol{x})$ of the DNN $f$ no matter what image $\boldsymbol{x}$ is used as input, i.e. $ \forall \boldsymbol{x} \in X, f(\boldsymbol{\delta})=f(\boldsymbol{\delta}+\boldsymbol{x}),$ where $X$ can be any unlabeled natural image set with or without intersection on the target DNN's training set. We keep the output feature of DNN unchanged by solving the optimization problem as
\begin{equation}\label{eq:costfunction}
\mathop{\text{maximize}}\limits_{\|\boldsymbol{\delta}\|_p\le \xi} \mathop{\mathbb{E}}\limits_{\boldsymbol{x}\in X}
\left(\mathcal{L}(f(\boldsymbol{\delta}),f(\boldsymbol{\delta}+\boldsymbol{x}))\right),    
\end{equation}
where $\xi$ is a small value ensuring the quasi-imperceptibility of $\boldsymbol{\delta}$, and $\mathcal{L}$ is a loss function used to measure the similarity between vectors as discussed in detail later. 

Since the constraint function $\|\boldsymbol{\delta}\|_p\leq \xi$ can be easily realized by clamping the variable's value during training, the key to the algorithm falls into how to quantify the similarity between $f(\boldsymbol{\delta}) $ and $f(\boldsymbol{\delta} +\boldsymbol{x})$. It is worth noting that feature $f(\cdot)$ is a high-dimensional vector. Typically, there are several metrics to measure the similarity between vectors, such as cosine similarity, Euclidean distance, and Kullback–Leibler divergence, which are defined as follows.

\begin{enumerate}
    \item \textbf{Cosine Similarity Loss (CoSLoss).} 
    Cosine similarity is the cosine of the angle between vectors. It treats the input vectors $\boldsymbol{a},\boldsymbol{b}$ as normalized, which is a judgment of orientation and not magnitude, and its outcome is bounded in $[0,1]$. 
    The higher the cosine similarity is, the more similar the two vectors are. Therefore, the CoSLoss can be defined as
    \begin{equation}
    \mathcal{L}_{\mathrm{CoS}}(\boldsymbol{a}, \boldsymbol{b})
    = 1-\frac{\boldsymbol{a}\cdot \boldsymbol{b}}{\|\boldsymbol{a}\|\| \boldsymbol{b}\|}
    = 1-\frac{\sum\limits_{i=1}^{n}a_ib_i}
    {\sqrt{\sum\limits_{i=1}^{n}a_i^2}\sqrt{\sum\limits_{i=1}^{n}b_i^2}},
    \label{equ:CoSLoss}
    \end{equation}
    where $n$ denotes the dimension of $\boldsymbol{a}$ and $\boldsymbol{b}$, $a_i$ and $b_i$ denote the $i$-th dimension of the vector $\boldsymbol{a}$ and $\boldsymbol{b}$, respectively.
    
    \item \textbf{Euclidean Distance Loss (EDLoss).} Euclidean distance is defined as the length of the line segment between the two points. The loss according to this similarity metric can be directly designed as
    \begin{equation}
    \mathcal{L}_{\mathrm{ED}}(\boldsymbol{a}, \boldsymbol{b}) = \|\boldsymbol{a}-\boldsymbol{b}\|=\sqrt{\sum\limits_{i=1}^n (a_i-b_i)^2}.
    \label{equ:csloss}
    \end{equation} 
    
    \item \textbf{Kullback–Leibler Divergence Loss (KLDLoss).} If we treat the input vectors  $\boldsymbol{a}$ and $\boldsymbol{b}$ as two probability distributions, then Kullback–Leibler divergence is defined to measure the similarity between two probability distributions. 
    KLD is an asymmetric measurement function used to measure how well a probability distribution $\boldsymbol{a}$ fits the target distribution $\boldsymbol{b}$.
    In our work, we regard $f(\boldsymbol{\delta})$ as the target distribution, since $\boldsymbol{\delta}$ is the main part we optimized.
    \begin{equation}
    \label{eq:KL}
    \mathcal{L}_{\mathrm{KLD}}(\boldsymbol{a},\boldsymbol{b})=\text{KLD}(\boldsymbol{a},\boldsymbol{b})=\sum\limits_{i=1}^n b_i \log \frac{a_i}{b_i}.
    \end{equation}
\end{enumerate}

\begin{algorithm}[b]
\caption{Finding dominant patterns for a given DNN}
\label{ag:dp}
\textbf{Input}: DNN $f$,  loss function $\mathcal{L}_{\mathrm{CoS}}$, training set $X$, perturbation magnitude $\xi$, batch size $b$, maximum number of epochs $m$, learning rate for Adam optimizer $lr$.\\
\textbf{Output}: dominant pattern $\boldsymbol{\delta}$ 
\begin{algorithmic}[1]
\STATE Initialize $\boldsymbol{\delta}\leftarrow 0, t \leftarrow 0$.
\WHILE{$t<m$}
    \FOR{each batch of data $B_i \subset X:len(B_i)=b$}
        \STATE $\boldsymbol{g} \leftarrow \underset{\boldsymbol{x}\sim B_i}{\mathbb{E}}\left[\nabla_{\boldsymbol{\delta}}\mathcal{L}_{\mathrm{CoS}}(f(\boldsymbol{\delta}), f(\boldsymbol{\delta}+\boldsymbol{x}))\right]$
        \STATE $\boldsymbol{\delta} \leftarrow \boldsymbol{\delta} + \Gamma (\boldsymbol{g},lr)$  ~~~~\text{\# the Adam optimizer}
        \STATE $\boldsymbol{\delta} \leftarrow \min(\max(\boldsymbol{\delta}, -\xi), \xi)$ ~~~~\text{\# clamp the pattern}
    \ENDFOR
    \STATE $t \leftarrow t+1$
\ENDWHILE
\end{algorithmic}
\end{algorithm}


Theoretically, CoSLoss is a better choice, because EDLoss is easily affected by magnitude, while KLDLoss is an asymmetric metric thus is not easy to optimize.
To verify our choice, we choose a classification task to compare the impact of different losses on feature similarity.
We define a metric called \emph{dominance ratio}, i.e. the ratio of perturbed images ($\boldsymbol{\delta}+\boldsymbol{x}$) that are predicted as the dominant pattern's class, to measure the performance of these losses.
We attack VGG19 \cite{vgg} on ImageNet with three mentioned loss, which is calculated at the logit layer of DNNs since it is highly related to the dominance ratio.
In Table \ref{tab:loss}, we list the dominance ratio of patterns found by minimizing different losses (the detailed setting could be found in Section \ref{sec:exp}). Because cosine similarity outperforms the others, we choose it as the objective in the algorithm. 

\begin{table}[h]
\centering
\caption{Dominance of dominant patterns for VGG19 with different similarity metrics}
\begin{tabular}{@{}ccccc@{}}
\toprule
 loss& CoSLoss & EDLoss & KLLoss \\ \midrule
dominance ratio (\%)    & 92.93  & 30.40     & 9.24                               \\ \bottomrule
\end{tabular}
\label{tab:loss}
\end{table}

The optimization is iteratively proceeded over training data to minimize the CoSLoss. Adam \cite{kingma2014adam} optimizer is applied and the perturbation is clamped by $\ell_\infty$ norm in each iteration to satisfy the magnitude requirement. The details are summarized in Algorithm \ref{ag:dp}.


\section{Experiments}\label{sec:exp}
In this section, we first verify the existence of dominant patterns, showing that the proposed algorithm can indeed find patterns dominating DNNs. Further, the layer-wise output of a DNN is visualized to better understand the dominance process.
After that, we demonstrate that dominant patterns can be applied to UAP tasks and defeat state-of-the-art methods. Moreover, we discuss its superiority over UAP methods in terms of their dominance and task transferability.

\subsection{Setup}\label{sec:setup}
To verify the existence and performance of dominant patterns, several representative DNNs are chosen as target models, including AlexNet \cite{alexnet}, GoogLeNet \cite{googlenet}, VGG16 \cite{vgg}, VGG19 \cite{vgg}, ResNet50 \cite{resnet}, ResNet152 \cite{resnet}, and DenseNet121 \cite{densenet} (hereinafter abbreviated as AN, GLN, VGG16, VGG19, RN50, RN152, and DN121, respectively).  The target models we used are got from Torchvision \cite{paszke2019pytorch} and are pre-trained on ImageNet \cite{imagenet}. The hyper-parameters in Algorithm \ref{ag:dp} are set as $b=32$, $m=10$, $lr=0.01$. All experiments are performed on PyTorch \cite{paszke2019pytorch} with NVIDIA GeForce RTX 2080Ti GPUs.

In our experiment, the public datasets we used include validation set of ImageNet ILSVRC2012 \cite{imagenet} ($50,000$ images), MSCOCO 2017 \cite{coco} ($118,287$ images) and CIFAR10 \cite{cifar} ($50,000$ images). Among them, the ImageNet validation set is not divided into training and test sets, therefore we randomly split it into two parts, where $40,000$ pieces are used for training, and the rest $10,000$ are used for testing. We also create an artificial dataset with $4,000$ images in four classes ($1,000$ per class) by using BigGAN \cite{biggan} of which the official model can only generate four classes.


In UAP tasks, comprehensive baselines are chosen for comparsion, including UAP \cite{uap}, GAP \cite{gap}, NAG \cite{nag}, FT-UAP \cite{ftuap}, TUAP \cite{tuap}, FFF \cite{fff}, AAA \cite{aaa}, GD-UAP \cite{gduap}, and PD-UA \cite{pdua}. For fair comparison, we inherit all hyperparameter settings from the respective papers, and the generated dominant patterns and UAPs are restricted with $\|\boldsymbol{\delta}\|_\infty \le 10$ for images in range [0, 255]. To evaluate fooling performance, the most common metric \emph{Fooling Rate} is used, which denotes the proportion of test images for which the prediction of the target DNN is changed after adding the perturbation.

\begin{figure}[b]
\vspace{-5mm}
\centering
\subfigure[VGG16 (ImageNet)]{
\includegraphics[width=0.22\textwidth]{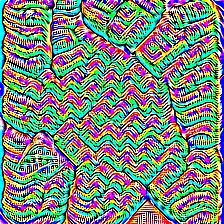}}
\hspace{0.05in}
\subfigure[VGG19 (ImageNet)]{
\includegraphics[width=0.22\textwidth]{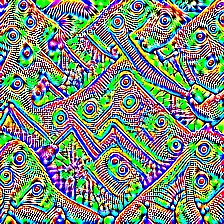}}
\hspace{0.05in}
\subfigure[RN152 (ImageNet)]{
\includegraphics[width=0.22\textwidth]{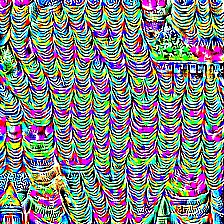}}
\hspace{0.05in}
\subfigure[DN121 (ImageNet)]{
\includegraphics[width=0.22\textwidth]{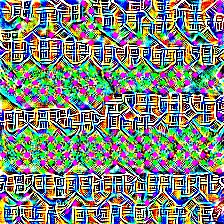}}\\
\vspace{-2mm}\hspace{0.3mm}
\subfigure[VGG16 (COCO)]{
\includegraphics[width=0.22\textwidth]{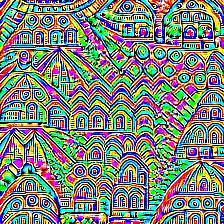}}
\hspace{0.05in}
\subfigure[VGG19 (COCO)]{
\includegraphics[width=0.22\textwidth]{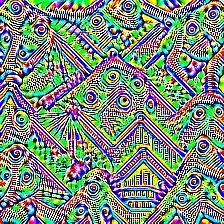}}
\hspace{0.05in}
\subfigure[RN152 (COCO)]{
\includegraphics[width=0.22\textwidth]{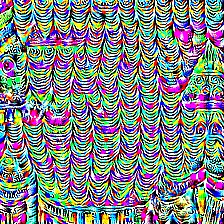}}
\hspace{0.05in}
\subfigure[DN121 (COCO)]{
\includegraphics[width=0.22\textwidth]{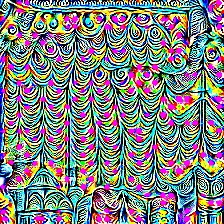}}
\caption{Visualization of dominant patterns for four representative DNNs.}
\label{fig:dp} 
\end{figure}

\subsection{Dominant patterns}
We first implement Algorithm 1 on four representative DNNs with ImageNet validation set and COCO 2017 training set. The calculated dominant patterns are shown in Figure \ref{fig:dp}.
Since the max magnitudes of the results are restricted to $\xi=10$ which is nearly invisible, we map the magnitudes from [-10,10] to [0,255] for better observation. 

\begin{figure}[b]
    \centering
    \vspace{-5mm}
    \includegraphics[width=.95\textwidth]{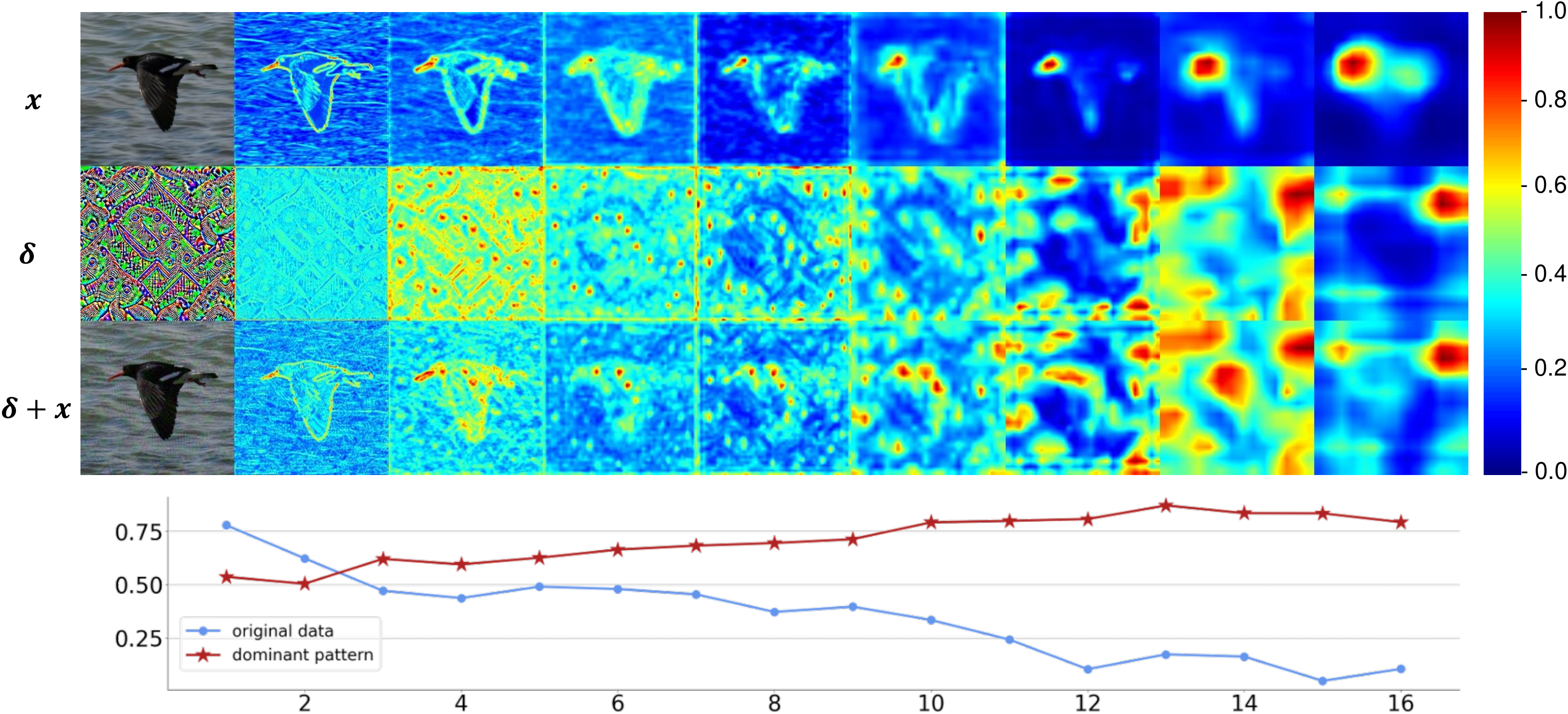}
    \caption{A visualization of the output of all ReLU layers in VGG19. The upper part is the pseudo-color image of even layers' normalized output. The lower part shows the changes in the SSIM correlation between the adversarial example and the original image (blue) / dominant pattern (red).}
    \label{fig:exp4-1}
\end{figure}

Visually, the found patterns are regular textures, which contain certain information. According to Table \ref{tab:dp}, although such a pattern is meaningless compared with natural images, the corresponding DNN identifies it as a certain class (named \emph{dominant class}) with very high confidence. We repeat the experiments several times and find a stable dominant class for each setting. 
Quantitatively, their dominance ratio (defined in Section \ref{sec:method}) are also reported. After adding a dominant pattern, more than three-quarters of images are misled to the dominant class, which originally belong to 1,000 classes. The high dominance ratio demonstrates that the generated dominant patterns successfully attract DNN's attention and take control of its output. 



\begin{table}[t]
\centering
\caption{Information of dominant patterns in Figure \ref{fig:dp}.}
\resizebox{\linewidth}{!}{
\begin{tabular}{c|c|ccccc}
\toprule
\multirow{2}{*}{Training Set } & \multirow{2}{*}{Metric}& \multicolumn{4}{c}{Target Net} \\
& &  VGG16& VGG19   & RN152 & DN121 \\ \midrule
\multirow{3}{*}{ImageNet} & dominant class & sock  & common iguana     & theater curtain      & carton                          \\
&confidence (\%)     & 99.95  & 100.00     & 99.45     & 93.93                         \\
&dominance ratio (\%)    & 84.11  & 92.93      & 84.40     & 77.51                              \\ \midrule

\multirow{3}{*}{COCO} & dominant class & church  & common iguana    & theater curtain    & theater curtain  \\
&confidence (\%)    & 99.67	&100.00&	99.58 &	99.15 \\
&dominance ratio (\%) & 85.18&	87.89&	84.31&	74.30 \\ \bottomrule

\end{tabular}}
\label{tab:dp}
\vspace{-3mm}
\end{table}

By taking a closer look at the respective performance of dominant patterns trained on ImageNet and COCO, we find that for VGG19 and RN152, their dominant classes have not changed when given different data prior information. However, that is not true for the other two DNNs. Actually, the training procedure of dominant patterns depends on training data. Since the distribution of training set varies from each other, dominant patterns trained with them may belong to different classes. 


To investigate how these seemingly meaningless patterns dominate the DNNs, we visualize the layer-wise output of VGG19 corresponding to a randomly chosen natural image $\boldsymbol{x}$, the dominant pattern $\boldsymbol{\delta}$, and the perturbed image $\boldsymbol{\delta}+\boldsymbol{x}$ in Figure \ref{fig:exp4-1}. The horizontal axes denotes the ReLU layer index. Obviously, along with the forward propagation from input to output, the dominant pattern embedded in $\boldsymbol{\delta}+\boldsymbol{x}$ gradually seize the control role such that the features become less related to $\boldsymbol{x}$ and more similar to $\boldsymbol{\delta}$. As a numerical comparison, the Structural SIMilarity \cite{wang2004image} (SSIM) of $\boldsymbol{\delta} + \boldsymbol{x}$ v.s. $\boldsymbol{\delta}$ and that of $\boldsymbol{\delta} + \boldsymbol{x}$ v.s. $\boldsymbol{x}$ are given. 

\begin{figure}[t]
    \centering
    \includegraphics[width=.85\textwidth]{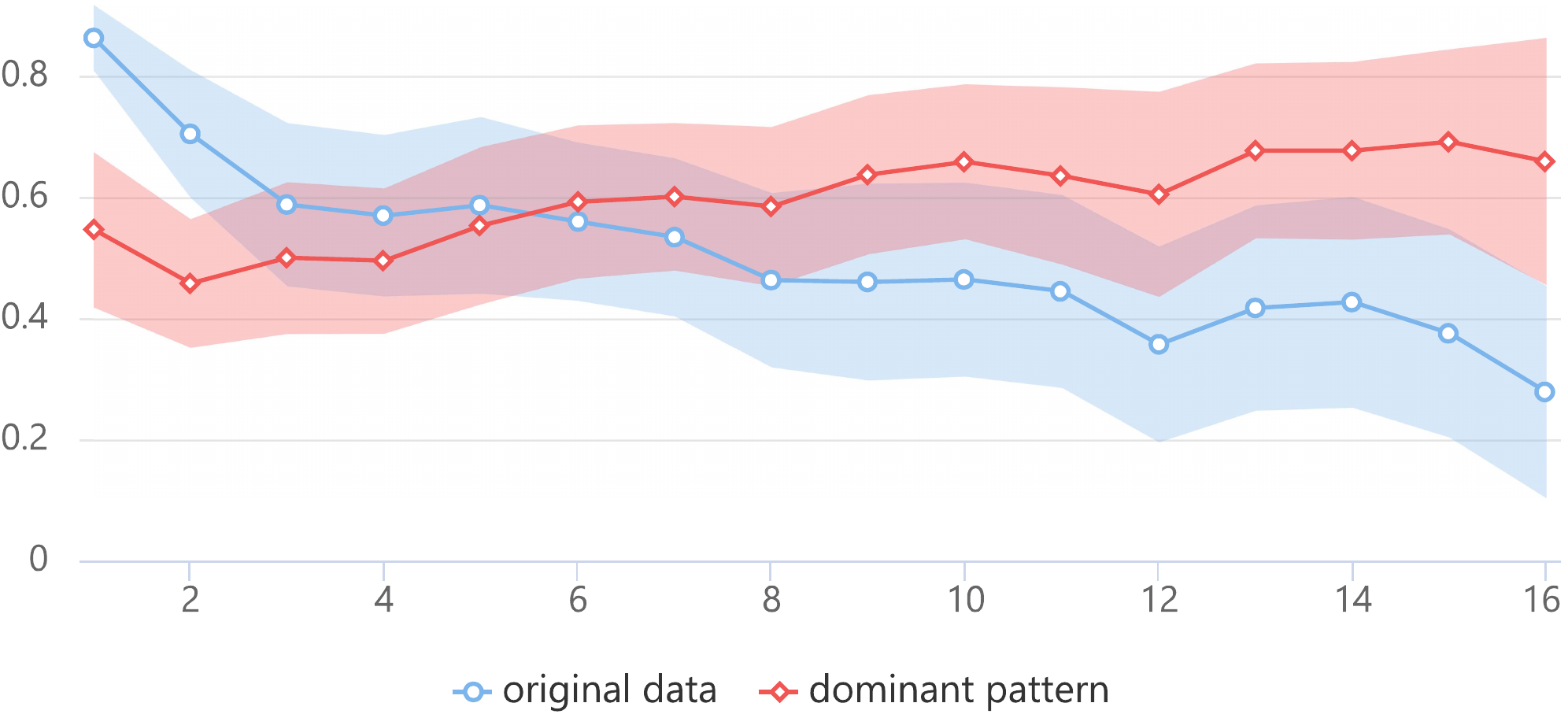}
    \caption{Average SSIM correlation between the adversarial examples and original data (blue) / dominant pattern (red) with their standard deviation (presented as arearange series). }
    \label{fig:exp4-2}
    \vspace{-5mm}
\end{figure}

The above is not a cherry-picked case. We repeat this experiment for all the 10,000 images in the test set and report the average SSIM with standard deviations in Figure \ref{fig:exp4-2}, from which one could observe a similar trend. This phenomenon indicates that the damage of dominant patterns permeates through the whole DNN, and has already been accomplished at the output of the backbone (the $16^{th}$ ReLU layer). That is to say, dominant patterns destroy DNNs by disabling their ability to extract meaningful features from natural images. Since feature extraction is a key in the computer vision, this inspires us that dominant patterns are attributes for DNNs and can serve as an all-in-one perturbation for DNNs with the same backbones in different tasks.


\subsection{Performance in UAP task}
\label{sec:uap}
 
In a universal attack task, one seeks a universal perturbation $\boldsymbol{\delta}$ that can change the prediction of most images from dataset $X$, i.e., the attackers find a $\|\boldsymbol{\delta}\|_p\le \xi$ that satisfies
\begin{equation}\label{eq:uap}
    f(\boldsymbol{x}+\boldsymbol{\delta})\neq f(\boldsymbol{x})~ \text{for most}~ \boldsymbol{x}\in X
\end{equation}
Compare this equation with the dominant pattern's definition in equation (\ref{eq:concept}), it can be easily found that the dominant patterns can also change DNN's prediction. Therefore, they can be applied to UAP tasks. We discuss its performance from two aspects according to the type of training set used.

\begin{figure}[t]
    \centering
    \includegraphics[width=.9\textwidth]{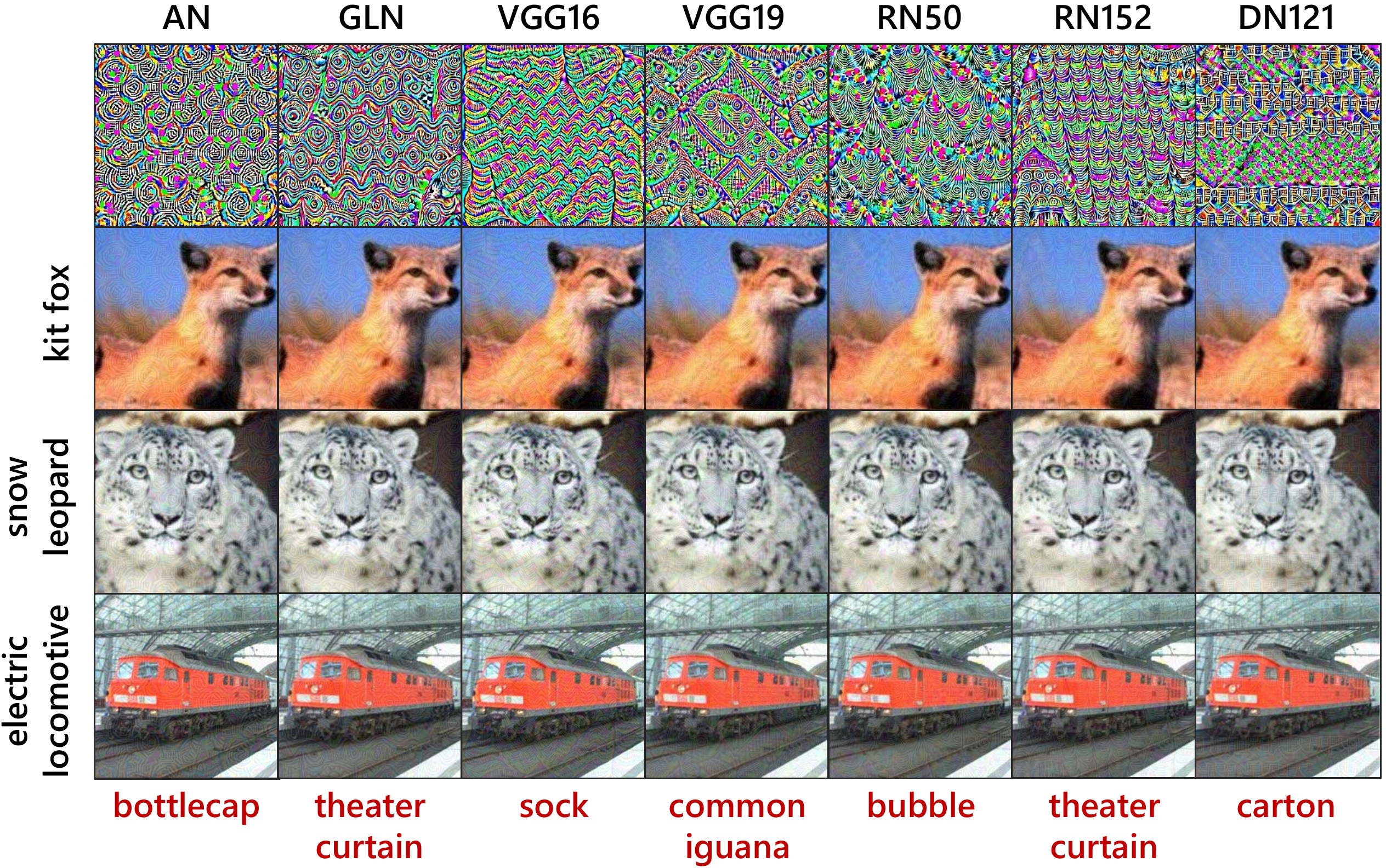}
    \caption{Some adversarial examples generated from dominant patterns. The text at the bottom denotes the predicted classes for dominant patterns and corresponding adversarial examples.}
    \label{fig:exp2}
    \vspace{-4mm}
\end{figure}

\paragraph{Train with original data.} The most attack-friendly situation is to train UAP on labeled data that is also used for training the DNNs, which is called label-dependent attack. In this case, UAPs could achieve high fooling performance. The fooling rates calculated by applying different attack methods to different DNNs are shown in Table \ref{tab:uap}, where the best fooling rates are highlighted in bold, and the second-best ones are underlined. For comparative experiments whose results are not provided in the corresponding papers, we mark them with '-'. From the table, it can be concluded that our dominant patterns defeat the SOTA method.
Figure \ref{fig:exp2} shows some adversarial examples generated from dominant patterns for different DNNs. 

\begin{table}[h]
\centering
\caption{Fooling Rates (\%) of dominant patterns (trained with original dataset) and label-dependent UAP methods.}
\resizebox{\textwidth}{!}{
\begin{tabular}{@{}c|cccccccc@{}}
\toprule
Attack type&Method & AN & GLN & VGG16         & VGG19        & RN50      & RN152     & DN121 \\ \midrule
\multirow{6}{*}{\tabincell{c}{Label-\\dependent}}&UAP \cite{uap}    & 93.30                        & 78.90                          & 78.30           & 77.80          & -              & 84.00             & 77.38                            \\
&GAP \cite{gap}    & -                           & 82.70                          & 83.70           & 80.10          & 62.83          & 59.19          & 71.13                            \\
&NAG \cite{nag}    & \underline{96.44}                       & \underline{90.37}                         & 78.30           & 83.78         & 86.64          & 87.24          & -                                \\
&FT-UAP \cite{ftuap} & -                           & 85.80                          & 93.50           & 94.50          & \underline{93.60}           & \underline{92.70}           & -                                \\
&TUAP \cite{tuap}   & 96.17                       & 88.94                         & \underline{94.30}           & \underline{94.98}         & -              & 90.08          & \underline{86.45}                            \\
&Ours (ImageNet)   & \textbf{96.79}              & \textbf{90.45}                & \textbf{97.40} & \textbf{97.45} & \textbf{94.38} & \textbf{94.17} & \textbf{95.61}                   \\ \bottomrule
\end{tabular}}
\label{tab:uap}
\vspace{-2mm}
\end{table}

\paragraph{Train with proxy data.} 
Although label-dependent attacks can achieve high fooling performance, the task and training data of the target DNN is generally unavailable to attackers. In this case, we can use an unlabeled proxy dataset that is irrelevant to target models. Such data can provide data prior information, leading to label-free attacks. Some attackers also attempt to conduct data-free attacks, where no data is utilized during training. 

According to Algorithm \ref{ag:dp}, finding dominant patterns is a self-supervised process thus does not requires any data labels, which satisfies a label-free setting. We use MSCOCO \cite{coco} and artificial images generated by BigGAN \cite{biggan} as our proxy dataset. Details of these datasets are stated in Section \ref{sec:setup}. The fooling rates are reported in Table \ref{tab:2}, where the performance of data-free UAPs (row 1-4) and label-free UAPs (row 5-7) are given for comparison. 
Combined with Table \ref{tab:uap}, it can be concluded that dominant patterns obtained from proxy data not only outperform all label-free and data-free attacks, but also are comparable to SOTA label-dependent attacks. The outstanding performance further enhances the practicality of dominant patterns. 
Notice that, the performance gap between COCO and BigGAN is mainly due to the lack of diversity in BigGAN pictures (with only four classes).

 
\begin{table}[htbp]
\centering
\caption{Fooling Rates (\%) of dominant patterns (trained with proxy dataset) and data-free or label-free UAP methods.}
\resizebox{\textwidth}{!}{
\begin{tabular}{@{}c|ccccccc@{}}
\toprule
Attack type & Method       & AN        & GLN      & VGG16         & VGG19         & RN50      & RN152    \\ \midrule
\multirow{4}{*}{\tabincell{c}{Data-free}}&FFF \cite{fff}         & 80.92          & 56.44          & 47.10           & 43.62          & -              & -              \\
&AAA \cite{aaa}         & 89.04          & 76.80           & 71.59          & 72.84          & -              & 60.72          \\

&PD-UA \cite{pdua}       & -              & 67.12          & 53.09          & 48.95          & 65.84          & 53.51          \\ 
&GD-UAP (NP) \cite{gduap}      & 84.88           & 58.62          & 45.47          & 40.68          & -              & 29.78           \\\midrule
\multirow{5}{*}{\tabincell{c}{Label-free}}& GD-UAP (DP) \cite{gduap}      & 91.54           & 83.54          & 77.77         & 75.51          & -              & 66.68           \\
&TUAP (COCO) \cite{tuap} & 89.90 & 76.80 & 92.20 & 91.60 & - & 79.90\\ 
&TUAP (BigGAN)  \cite{tuap} & 88.51 & 75.32 & 92.30 & 92.47 &- & 79.24 \\
&Ours (COCO) & \textbf{96.49} & \textbf{90.25} & \textbf{97.44} & \textbf{97.30} & \textbf{94.44} & \textbf{93.45}          \\
&Ours (BigGAN)  & \underline{93.61}          & \underline{86.65}           & \underline{95.44}          & \underline{94.78}          & \underline{91.69}          & \underline{82.62} \\ \bottomrule
\end{tabular}}
\label{tab:2}
\vspace{-3mm}
\end{table}



\subsection{Dominant patterns beyond UAP}


In previous subsections, we demonstrate the existence of dominant patterns and their outstanding performance in UAP tasks. 
Dominant patterns are far from UAPs, which can be illustrated from two following aspects:

\paragraph{Dominant patterns can be UAPs, while UAPs are not dominant patterns.}
A UAP pursues to change the prediction output, while a dominant pattern pursues to keep the output feature unchanged with different inputs. 
Dominant patterns can be used as UAPs with a high fooling rate as shown in Section \ref{sec:uap}, but UAPs cannot achieve the same dominance ratio with dominant patterns. To show that, we report the top-k ratios obtained by UAPs and dominant patterns in Table \ref{tab:topk vs other uap}. 
Here, the top-k ratio means the overall proportion of top k most frequently occurred classes. 



Although it has been discovered in \cite{uap} and \cite{dominant} that UAPs tend to aggregate images to some dominant labels, dominant patterns gather perturbed images to several certain classes to a large extent that most UAPs cannot achieve, as shown in Table \ref{tab:topk vs other uap}. This verifies our claim that "UAPs are not dominant patterns". The direction aggregation implies that DNNs' vulnerability is class-dependent and there are certain flaws that are extremely fragile.
TUAP is a newly proposed method that can take into account the multiple dimensions of the logit vector to fully utilize information. It satisfies some of the conditions for obtaining dominant patterns, so that it can perform well on dominance ratio.

\begin{table}[t]
\centering
\caption{top-k ratios (\%) of some universal attack methods VGG19 (trained with ImageNet).}
\begin{tabular}{@{}ccccccc@{}}
\toprule
top-k ratio & UAP   & GAP   & NAG  & TUAP  & Ours            \\ \midrule
top-1   & 15.67 & 8.09  & 1.59 & 87.30 & \textbf{92.93} \\
top-3   & 30.83 & 18.21 & 3.78 & 88.26 & \textbf{95.07} \\
top-5   & 35.01 & 25.99 & 5.85 & 88.94 & \textbf{95.51} \\ \bottomrule
\end{tabular}
\label{tab:topk vs other uap}
\end{table}

\paragraph{Dominant patterns can be transferred to downstream tasks.}
According to Figure \ref{fig:exp4-1} and dominant patterns' definition, even the features extracted by the backbone of DNNs can also be controlled by dominant patterns.
Theoretically, this phenomenon will be kept if the backbone of a pre-trained DNN remains unchanged in the downstream tasks.
However, when using the pre-trained network, it is common for trainers to freeze the parameters of the backbone and then train a head for downstream tasks. This is a potential risk and can be exploited by our dominant patterns.

To verify this assumption, we transfer the VGG19, which pre-trained on ImageNet, to CIFAR10 classification task by freezing the parameters in convolution layers (backbone) and fine-tune the ones in fully connected layers (classifier head). 
And then we use the dominant pattern obtained from the pre-trained VGG19 to attack the new DNN fine-tuned in CIFAR10. Consequently, the accuracy in CIFAR10 task is promoted to 87.02\%, and the results are shown in Table \ref{tab:cifar}.

\begin{table}[htbp]
\centering
\caption{Performance of dominant pattern (transferred from ImageNet) on CIFAR10.}
\resizebox{\textwidth}{!}{
\begin{tabular}{@{}ccccc@{}}
\toprule
class name  & confidence (\%)  & fooling rate (\%)  & dominance ratio (\%)  
& CoSLoss (backbone feature) \\ \midrule
frog  & $100.0$ & $89.47$ &  $99.99$ 
& $5.089\times10^{-2}$ \\ \bottomrule
\end{tabular}}
\label{tab:cifar}
\end{table}

The dominant pattern still serves as a strong perturbation which shows extremely high confidence and dominance ratio.
Note that since CIFAR10 is a 10-category classification task, the theoretical highest fooling rate is 90\% because the clean samples which share the same prediction with the dominant pattern will not change their predictions after attacked. 

More importantly, the reported CoSLoss remains at a very low level on the fine-tuned network. This indicates that the DNN is still in control of the dominant pattern, even though the task has already changed.
This phenomenon further verifies that the proposed dominant pattern can attack any DNNs that use the same pre-trained backbone without any changes. It is also the reason why we call the dominant pattern "the unique nature of the DNN".

\section{Conclusion}\label{sec:conclusion}
In this work, we provide a new perspective on perturbations, and propose that there are dominant patterns hidden in DNNs, which is a critical feature of DNN and can help understand and improve the DNN's robustness. Although small and invisible in the image space, dominant patterns have some meaningful patterns that have the ability to mislead the majority of natural images to their own classes. Through numerical experiments, we verify the existence of such pattern and visualize how its dominance takes place. Further, we demonstrate the outstanding performance of dominant patterns in UAP tasks, and analyze their particularity beyond the traditional attacks. In the future, it is possible for researchers to go deeper into the properties of dominant patterns, and apply it to realize better model defense in further study. Our proposal reveals the underlying threat hidden in DNNs, but since we focus only on the digital space and white-box attacks, it does not induce imminent threats to the applications of DNNs in the physical world.

\newpage
\appendix
\setcounter{figure}{0}
\setcounter{table}{0}
\makeatletter 
\renewcommand{\thefigure}{A.\@arabic\c@figure}
\renewcommand{\thetable}{A.\@arabic\c@table}
\makeatother
\section{Error bars for experiment on Universal Adversarial Perturbation task}
\label{appendix:errorbar}
\begin{table}[htbp]
\centering
\caption{Fooling rates(\%) of dominant patterns on the ImageNet classification task.}
\resizebox{\linewidth}{!}{
\begin{tabular}{@{}cccccccc@{}}
\toprule
Method & AN & GLN & VGG16         & VGG19        & RN50      & RN152     & DN121 \\ \midrule
Ours   & 96.79$\pm$0.09              & 90.45$\pm$0.17               & 97.40$\pm$0.08 & 97.45$\pm$0.06 & 94.38$\pm$0.22 & 94.17$\pm$0.67 &95.61$\pm$0.45                   \\ 
Ours (COCO) & 96.49$\pm$0.10 & 90.25$\pm$0.12 & 97.44$\pm$0.22 & 97.30$\pm$0.13 & 94.44$\pm$0.29 & 93.45$\pm$1.66 & 91.14$\pm$0.12 \\
Ours (BigGAN) & 93.61$\pm$0.33 & 86.65$\pm$0.36 & 95.44$\pm$3.15 & 94.78$\pm$0.06 & 91.69$\pm$0.24 & 82.62$\pm$0.98 & 80.61$\pm$1.73\\
\bottomrule
\end{tabular}}
\label{tab:uap_error}
\end{table}

\section{Dominant patterns found by utilizing fake dataset}
\begin{figure}[H]
\centering
\subfigure[VGG16 (BigGAN)]{
\includegraphics[width=0.22\textwidth]{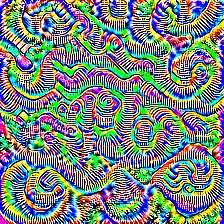}}
\hspace{0.05in}
\subfigure[VGG19 (BigGAN)]{
\includegraphics[width=0.22\textwidth]{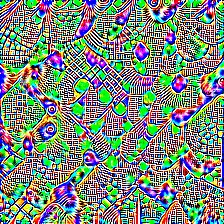}}
\hspace{0.05in}
\subfigure[RN152 (BigGAN)]{
\includegraphics[width=0.22\textwidth]{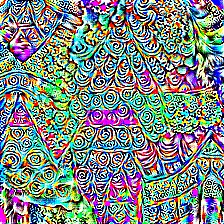}}
\hspace{0.05in}
\subfigure[DN121 (BigGAN)]{
\includegraphics[width=0.22\textwidth]{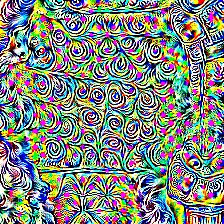}}
\caption{Visualization of dominant patterns for four representative DNNs.}
\label{fig:dp_biggan} 
\end{figure}

Here we post in Figure \ref{fig:dp_biggan} the dominant patterns for four DNNs trained with artificial images generated from BigGAN \cite{biggan}, with their information in Table \ref{tab:dp_biggan}. 

\begin{table}[H]
\centering
\caption{Information of dominant patterns in Figure \ref{fig:dp_biggan}.}
\begin{tabular}{@{}ccccc@{}}
\toprule
target net& VGG16 & VGG19  & RN152 & DN121  \\ \midrule
dominant class & brain coral  & spider monkey    & vestment      & borzoi                          \\
confidence (\%) &100.00&	93.98&	98.54&	58.05\\
dominance ratio (\%)&85.78 &	55.27&	62.08&	28.12\\
\bottomrule
\end{tabular}
\label{tab:dp_biggan}
\end{table}

The confidence and dominance ratio are not so satisfying as the ones trained with ImageNet and COCO dataset, especially for VGG19 and DN121. We think this is a foreseeable result, considering that the BigGAN based dataset is not exactly a natural image set, but close to the natural image distribution. More importantly, due to the restriction of open source, the generated images can only cover 4 over 1,000 classes in ImageNet, which makes it a rather challenging task.
\newpage
\section{Adversarial Examples generated by utilizing proxy image set.}
\begin{figure}[htbp]
    \centering
    \includegraphics[width=.85\textwidth]{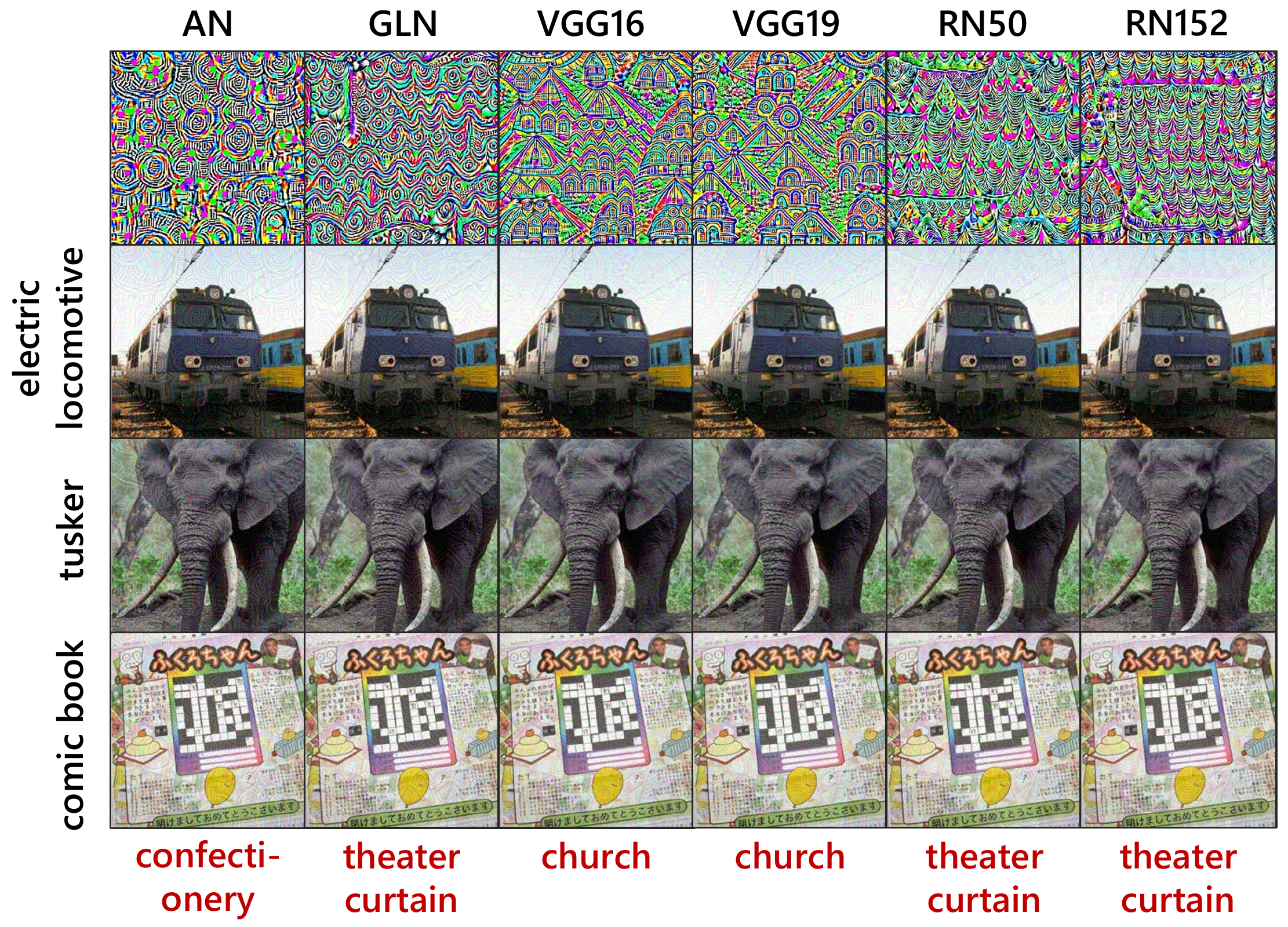}
    \caption{Adversarial examples generated from dominant patterns (trained on COCO).}
    \label{fig:ae_coco}
\end{figure}

\begin{figure}[htbp]
    \centering
    \includegraphics[width=.85\textwidth]{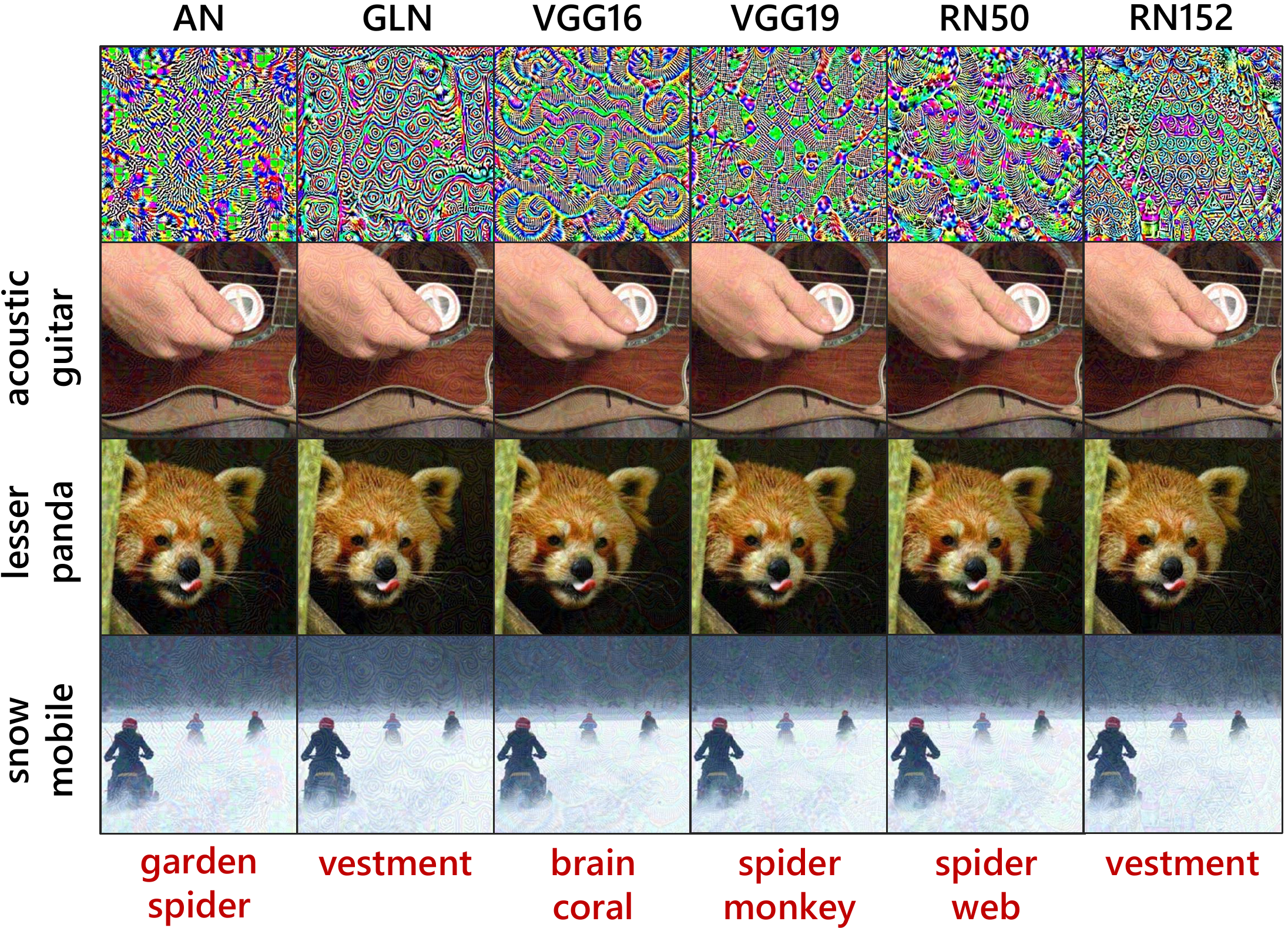}
    \caption{Adversarial examples generated from dominant patterns (trained on BigGAN dataset).}
    \label{fig:ae_biggan}
\end{figure}

Figure \ref{fig:ae_coco} and Figure \ref{fig:ae_biggan} show some perturbed images attacked by dominant patterns that are trained with COCO and BigGAN dataset, respectively.
\end{document}